\documentclass[10pt,conference,a4paper]{IEEEtran}
\usepackage[utf8]{inputenc}
\usepackage{amssymb}

\ifCLASSINFOpdf
   \usepackage[pdftex]{graphicx}
  \graphicspath{{images/}}
\else
\fi
\usepackage{amsmath}
\usepackage{bbold}
\usepackage{algorithmicx}
\usepackage{algorithm}
\usepackage{algpseudocode}
\usepackage{booktabs}

\usepackage{array}
\usepackage{multirow}
\usepackage{multicol}

\ifCLASSOPTIONcompsoc
 \usepackage[caption=false,font=normalsize,labelfont=sf,textfont=sf]{subfig}
\else
 \usepackage[caption=false,font=footnotesize]{subfig}
\fi
\usepackage{url}

\usepackage[pagebackref=true,breaklinks=true,letterpaper=true,colorlinks,bookmarks=false]{hyperref}

\begin{document}
%
\title{Optimal Transport as a Defense Against Adversarial Attacks}

\author{\IEEEauthorblockN{Quentin Bouniot, Romaric Audigier, Angelique Loesch}
\IEEEauthorblockA{Université Paris-Saclay, CEA, List, \\ F-91120, Palaiseau, France \\
    \{quentin.bouniot, romaric.audigier, angelique.loesch\}@cea.fr\\}\\
}


%


\maketitle

    \begin{abstract}
    
    Deep learning classifiers are now known to have flaws in the representations of their class. Adversarial attacks can find a human-imperceptible perturbation for a given image that will mislead a trained model. The most effective methods to defend against such attacks trains on generated adversarial examples to learn their distribution. 

    Previous work aimed to align original and adversarial image representations in the same way as domain adaptation to improve robustness. Yet, they partially align the representations using approaches that do not reflect the geometry of space and distribution.
    In addition, it is difficult to accurately compare robustness between defended models. Until now, they have been evaluated using a fixed perturbation size. However, defended models may react differently to variations of this perturbation size.
    
    In this paper, the analogy of domain adaptation is taken a step further by exploiting optimal transport theory. We propose to use a loss between distributions that faithfully reflect the ground distance. This leads to SAT (Sinkhorn Adversarial Training), a more robust defense against adversarial attacks.

    Then, we propose to quantify more precisely the robustness of a model to adversarial attacks over a wide range of perturbation sizes using a different metric, the Area Under the Accuracy Curve (AUAC). We perform extensive experiments on both CIFAR-10 and CIFAR-100 datasets and show that our defense is globally more robust than the state-of-the-art.

\end{abstract}



%
\IEEEpeerreviewmaketitle

\section{Introduction}

Deep learning models have become the state-of-the-art in many computer vision tasks. Yet, recent work \cite{szegedy_intriguing_2013}\cite{goodfellow_explaining_2014}\cite{papernot_limitations_2015} has shown that deep neural network can be easily fooled by \emph{adversarial examples}, a human-imperceptible noise added to an image. Misclassification of adversarial images shows that neural network still have issues generalizing their class representation. These adversarial attacks represent a major issue for the widespread adoption of deep learning in practical applications. 


Currently, training on adversarial examples generated is the most effective defense against adversarial attacks~\cite{goodfellow_explaining_2014}\cite{kurakin_adversarial_scale_2016}\cite{madry_towards_2017}, which means learning the distribution of adversarial examples.
This distribution is far from the original distribution as they lie in low probability regions~\cite{song2018pixeldefend}, \emph{i.e.} they are assigned low probability in the original distribution.
Defending against adversarial attacks can be seen as aligning the representations of adversarial and original images, similar to a domain adaptation problem~\cite{tramer_ensemble_2017}\cite{song_improving_2018}. Previous work aligns the moments of the distributions or explicitly constrain the cluster of classes~\cite{song_improving_2018}\cite{Mustafa_2019_ICCV}.
Yet, aligning the moments is an imprecise approach to define a distance. It reflects neither the geometry of space nor that of the distribution and can lead to instabilities during training.
As for the other case, enforcing images to their class clusters does not take into account the whole distribution.
These approaches result in a partial alignment of the representations.

Besides, defenses are tested against a fixed adversarial perturbation size to compare robustness to attacks. 
However, defended models may react differently to changes in the size of the perturbation. Some may be more robust at low perturbations and others at high perturbations, differences that will not be visible when evaluating with a fixed size.



In this article, we propose to directly constrain the representations of adversarial and original images to remain close after attack using a least-effort approach. To do this, we propose to use a discrepancy between distributions based on the theory of optimal transport. This allows us to consider the whole distribution with a loss that reflects geometric properties rather than an alignment of moments. We minimize the ground distance between representations to align and bring closer the distributions.

Then, we propose a metric to measure the robustness of a model to adversarial attacks. To do this, we use a wide range of sizes and take into account the evolution of performance over this range. Thus, we can compare the behaviour of different models for several perturbation sizes at the same time.

Our contributions can be summarized as follow : 
\begin{itemize}
   \item We propose \textbf{SAT} (Sinkhorn Adversarial Training), a new defense against adversarial attacks using a \textbf{Sinkhorn Divergence}, a loss based on the theory of optimal transport.
   \item We propose a \textbf{different evaluation metric for robustness} against adversarial examples: \textbf{Area Under the Accuracy Curve (AUAC)}.
\end{itemize} 

In the following, we introduce related works on adversarial attacks and defenses in Section~\ref{sec:related_work}. After preliminaries on the theory of optimal transport, we propose our defense in Section~\ref{sec:SAT} and our evaluation metric in Section~\ref{sec:auc}. Finally, we compare our defense with the state-of-the-art in Section~\ref{sec:exp}.




\section{Related Work} \label{sec:related_work}

In this section, we present an overview of previous work on adversarial attacks and defenses.

\subsection{Adversarial Attacks} 

After Szegedy \emph{et al.} \cite{szegedy_intriguing_2013} shed light on the effect of their attack on a standard classifier, several more efficient ways of computing adversarial examples have been discovered.
They can be grouped into three families:
\begin{itemize}
   \item \emph{Unbounded attacks}~\cite{carlini_towards_2017}\cite{Yao_2019_CVPR}\cite{8954314} solve a constrained optimization problem, and seek the smallest perturbation that will misclassify a given image. The perturbation size can be different for two images. These attacks are the most effective since the perturbation size is not bounded. Given enough iterations and time, they always find an adversarial example.
   \item \emph{Bounded attacks}~\cite{goodfellow_explaining_2014}\cite{kurakin_adversarial_scale_2016}\cite{madry_towards_2017}\cite{dong_boosting_2017} use the gradients of the model for a given image and perform several steps of projected gradient descent. The size of the perturbation found is fixed and the same for each images.
   \item \emph{Gradient Reconstitution attacks}~\cite{brendel_decision-based_2017}\cite{narodytska_simple_2017}\cite{papernot_limitations_2015} search the input space by following an approximation of the Jacobian or by a random walk. These greedy attacks are time consuming but they can bypass defenses based on obfuscating gradients~\cite{athalye_obfuscated_2018}. 
\end{itemize}

Having efficient attacks, \emph{i.e.} that find quickly good adversarial examples, is useful to improve the defense. Indeed, when generating adversarial examples for training, the use of efficient attacks leads to better examples and more robust models. Furthermore, efficient attacks provide an useful evaluation of defenses. 

To this end, we consider in our experiments the most efficient Bounded and Unbounded attacks, namely Projected Gradient Descent (PGD, a.k.a. Basic Iterative Method or Iterative Fast Gradient Sign Method)~\cite{kurakin_adversarial_scale_2016}\cite{madry_towards_2017} and Trust-Region Attack (TR)~\cite{Yao_2019_CVPR}.

\subsection{Adversarial Defenses}

There have been multiple attempts to protect against adversarial examples or to increase robustness of deep learning models.

\paragraph{Obfuscating gradients} A first type of approaches performs preprocessing~\cite{guo2018countering}\cite{liao_defense_2018}\cite{song2018pixeldefend} on images to detect or to mitigate the effect of the perturbation. Similarly, some methods complexify models~\cite{pangImprovingAdversarialRobustness2019a}\cite{s.2018stochastic} to make attacks more difficult to achieve. However, all these methods lead to obfuscated gradients and provide limited robustness. They are still vulnerable to more specific adversarial attacks~\cite{athalye_obfuscated_2018}\cite{Tramer_Carlini_Brendel_Madry_2020}.

\paragraph{Adversarial training}

As previously stated, the most successful defense against adversarial attacks is \emph{adversarial training}. Goodfellow \emph{et al.}~\cite{goodfellow_explaining_2014} considered a mixed batch of adversarial examples (generated during training) and original images whereas Madry \emph{et al.}~\cite{madry_towards_2017} proposed to use only adversarial examples in the batch. Tramèr \emph{et al.}~\cite{tramer_ensemble_2017} introduced several pre-trained models to generate different adversarial examples during training. Adversarial training aims to learn a single distribution containing both original and adversarial versions of training samples. 

In contrast, Song \emph{et al.}~\cite{song_improving_2018} framed the defense as a domain adaptation problem. They considered the original and adversarial training distributions as distinct and aimed to bring their representations in the logit space closer. They improved adversarial training by aligning the means with a Maximum Mean Discrepancy (MMD)~\cite{gretton_2007} norm and covariance with a Correlation Alignment (CORAL)~\cite{Sun_Saenko_2016}. Similarly, Mustafa \emph{et al.}~\cite{Mustafa_2019_ICCV} explicitly ensured a large distance between class prototypes and a small intra-class distance. 

Bringing the distributions closer in the logit space seems to have a significant effect for robustness. However aligning the moments of representations results in an unstable training and does not faithfully reflect the distance between the two distributions. In practice, this is reflected by artifacts (\emph{vanishing} or \emph{exploding gradients}) next to the extreme points of the distribution~\cite{feydy2018interpolating}. Constraining the distance between and within clusters of classes does not take into account the distribution of images as a whole.

Based on the theory of optimal transport, we propose to use Sinkhorn Divergences to consider the discrepancy between adversarial and original representations integrally and with more accurate geometric properties. We aim to minimize the ground distance between the representations.

\section{Sinkhorn Adversarial Training} \label{sec:SAT}

In this section, we describe our training protocol to increase robustness to adversarial examples. First, we introduce necessary background on Optimal Transport.

\subsection{Optimal Transport} 

The theory of Optimal Transport aims to find the minimal cost (in terms of distance) to move simultaneously several items (or a continuous distribution of items in the most extreme case) from one configuration onto another. In particular, it can be used for computing distances between probability distributions. 




In practice, solving the optimal transport problem is \emph{costly} and suffers from the \emph{curse of dimensionality}. Therefore, we consider its \emph{entropic regularization}~\cite{agenevay_phdthesis}\cite{Peyr__2019}:

\begin{align*}
 W_{2,\sigma}(\alpha, \beta) = &\min_{\pi \in \Pi(\alpha, \beta)} \int_{\Omega \times \Omega} ||x-y||_2^2d\pi(x,y) \\
 &+ \sigma \text{KL}(\pi | \alpha \otimes \beta)
\end{align*}

\noindent
with $\text{KL}(\pi | \alpha \otimes \beta) =  \int_{\Omega \times \Omega} \log \left(\frac{d\pi}{d \alpha d \beta} \right) d\pi$, the Kullback-Leibler divergence.
$\alpha$ and $\beta$ are two probability measures with finite second moment, $\Pi(\alpha,\beta)$ is the set of probability measures over the product set $\Omega \times \Omega$ with marginals $\alpha$ and $\beta$ and $\sigma > 0$ is the entropy regularization parameter. 
With this regularization, the problem can be solved efficiently on GPU~\cite{Cuturi_2013}.

Finally, \emph{Sinkhorn Divergences}~\cite{genevay2017learning}\cite{feydy2018interpolating} are defined as
$$
S_{\sigma}(\alpha,\beta) = W_{2,\sigma}(\alpha,\beta) - \frac{1}{2}W_{2,\sigma}(\alpha,\alpha) - \frac{1}{2}W_{2,\sigma}(\beta,\beta).
$$
\noindent
$S_{\sigma}$ interpolates between an \emph{Optimal Transport loss} and an \emph{MMD loss} depending on $\sigma$. The effect of the entropy will be studied in Section~\ref{sec:sigma_abl}. We will use this divergence to align the original and adversarial representations. 

\subsection{Training with a Sinkhorn Divergence}


Given a training set $\mathbf{D}^{tr} = \{x^{tr}_i\}$ with associated labels $\mathbf{Y}^{tr}$, adversarial examples $\hat{\mathbf{D}}^{tr} = \{\hat{x}^{tr}_i \}$ are generated to create a large shift in the model representations from a small perturbation. We aim to reduce this shift by using optimal transport. 

Our \emph{SAT} (Sinkhorn Adversarial Training) combines adversarial training with a Sinkhorn Divergence as a loss between the distributions of original and adversarial representations. We constrain not only the moments, but the totality of the distributions by making sure to respect the underlying geometrical properties in the representation space. We minimize the ground distance, \emph{i.e.} the $L_2$ distance, between representations of original images and their adversarial counterpart to bring them closer.


\begin{algorithm}[t]
    \caption{SAT}
    \label{alg:SAT_alg}
    \hspace*{\algorithmicindent} \textbf{Input :} Model $f$, training set $\mathbf{D}^{tr}$ and labels $\mathbf{Y}^{tr}$, number of epochs $T$, sinkhorn entropy $\sigma$ \\
    \hspace*{\algorithmicindent} \textbf{Output :} Defended model $f$
    \begin{algorithmic}[1]
        \For{$t=1$ to $T$}
            \For{batch $\mathbf{D}^{tr}_b \in \mathbf{D}^{tr}$ and labels $\mathbf{Y}^{tr}_b \in \mathbf{Y}^{tr}$}
                \State Use the current state of $f$ to generate an adversarial batch of images $\hat{\mathbf{D}}^{tr}_b$;
                \State Compute $\mathcal{L}_{SAT_\sigma}(\mathbf{D}^{tr}_b, \hat{\mathbf{D}}^{tr}_b, \mathbf{Y}^{tr}_b, f)$ (Eq.~\ref{eq:loss_SAT}) and update parameters of $f$ by backpropagation;
            \EndFor
        \EndFor
    \end{algorithmic}
\end{algorithm}

The loss function used to train a defended classifier with SAT is

\begin{equation} \label{eq:loss_SAT}
\begin{split}
    \mathcal{L}_{SAT_{\sigma}}(\mathbf{D}^{tr},\mathbf{\hat{\mathbf{D}}}^{tr}, \mathbf{Y}^{tr}, f) = \mathcal{L}_{CE}(\mathbf{\hat{\mathbf{D}}}^{tr}, \mathbf{Y}^{tr}, f) + \\ S_\sigma(f(\mathbf{D}^{tr}), f(\mathbf{\hat{\mathbf{D}}}^{tr})),
\end{split}
\end{equation}{}

\noindent
with $\mathcal{L}_{CE}$ the cross entropy used for classification, $f(\mathbf{D}^{tr}) = \{f(x^{tr}_i)\}$ and $f(\hat{\mathbf{D}}^{tr}) = \{f(\hat{x}^{tr}_i)\}$ the representations of original and adversarial training images respectively.
The full training algorithm is described in Algorithm~\ref{alg:SAT_alg}.

During training, for each new batch of images, we generate adversarial examples by attacking the current state of the model being trained with a state-of-the-art adversarial attack (\emph{e.g.} PGD). Then, we use the original and adversarial batch of images as separate distributions to compute the loss $\mathcal{L}_{SAT_{\sigma}}$ using Equation~\ref{eq:loss_SAT}. The size of the perturbation $\epsilon_{train}$ used by the attack during training is fixed. 


Having defined our defense, we now want to evaluate its robustness against adversarial examples. 
We want to assess performance both on original data and on varying degrees of adversarial data (within the limit of the definition). Currently, defenses are evaluated against an attack with a given perturbation size $\epsilon$. In general, the perturbation size used for evaluation is the same as the one chosen for adversarial training. However, this protocol does not allow an overall estimation of the robustness of a model.

\section{Measuring robustness} \label{sec:auc}

The section below raises concerns on the current evaluation protocol against adversarial examples, and proposes a different and more accurate metric. 

\subsection{Defining perturbation size}

As pointed out in the introduction to this paper, performance of models significantly differs when the size of the perturbation $\epsilon$ varies.

Each attack has a different way to generate adversarial examples and target a suitable loss function $\mathcal{L}$ (\emph{e.g.} cross-entropy). The perturbation size $\epsilon$ can be variable between attacked images or not. It is defined as the difference for a given norm (usually $L_\infty$ or $L_2$) between an image $x$ with label $y$ and the corresponding adversarial example $\hat{x}$ found by a given attack: $\epsilon = || x - \hat{x}||$.

On the one hand, with a \emph{Bounded attack}, \emph{e.g.} PGD~\cite{kurakin_adversarial_scale_2016}\cite{madry_towards_2017}, the size of the perturbation is explicitly defined and remains constant between attacked images:
\begin{equation*}
\begin{split}
    \hat{x}_{0} &= x + \eta \\
    \hat{x}_{n+1} &= \Pi_{x}^{\epsilon}\left(\hat{x}_{n}+\alpha \cdot \operatorname{sign}(\nabla_{\hat{x}_{n}} \mathcal{L}(\hat{x}_{n}, y, f) \right) \\
\end{split}
\end{equation*}
\noindent
with $\epsilon$ the size of the perturbation, $\eta$ a random initial noise, $\hat{x}_{n}$ the resulting adversarial image after the $n$-th iteration and $\alpha$ the iteration step size.  
$\Pi_x^\epsilon$ is the clip function, which ensures that $||\hat{x}_{n} - x|| \leq \epsilon$ and that $\hat{x}_{n}$ is a valid image, \emph{i.e.} the pixels are in the range $[0,1]$. 

On the other hand, with other attacks, \emph{e.g.} TR attack~\cite{Yao_2019_CVPR}, the size of the perturbation can be variable between images. The attack find the smallest perturbation that fool the model for a given image:
\begin{equation*}
    \arg \min_{||\Delta x||} \arg \max \mathcal{L}(x + \Delta x, y, f)
\end{equation*}{}
\noindent
with $\Delta x$ the perturbation found for image $x$ and $\epsilon = ||\Delta x||$.

Some models are more robust to high perturbations and other to low perturbations, which makes comparison difficult. Choosing the most robust model involves a trade-off in terms of accuracy.

\subsection{A metric for robustness}
\label{subsec:def_metric}

We noted a recurring trend to consider a fixed perturbation size $\epsilon$ when comparing defenses~\cite{song2018pixeldefend}\cite{song_improving_2018}\cite{pangImprovingAdversarialRobustness2019a}\cite{Mustafa_2019_ICCV}. A perturbation size that is often identical to the one used to generate the adversarial examples during training. However, in a realistic scenario, the model can be attacked with a wide range of perturbation sizes. In addition, even if the perturbation is high and visible to a human, the class may still be recognized. In this case, for generalization purposes, the model should still have relatively good performance.
Therefore, we recommend evaluating over a wide range of perturbation sizes.

In binary classification problems, the Receiver Operating Characteristic (ROC) curve is computed depending on the discrimination threshold. Then models are evaluated using the Area Under the Curve (AUC) to take into account both accuracy and recall.
Similarly, we propose to compute the accuracy for a wide range of perturbation sizes and then consider the \emph{Area Under the Accuracy Curve (AUAC)}.

Given a testing set $\mathbf{D}^{ts} = \{x^{ts}_i\}$ with associated labels $\mathbf{Y}^{ts} = \{y^{ts}_i\}$, we note $\ell_f(x^{ts}_i)$ the label attributed to the test datum $x_i^{ts}$ by a given model $f$.  We define the AUAC up to $\epsilon_{max}$:

\begin{equation*}
\begin{split}
    Acc(f,\epsilon, \mathbf{D}^{ts}) = \frac{\mathbb{1}_{ \{ x_i^{ts} \in \mathbf{D}^{ts} | \ell_f(\hat{x}^{ts}_i) = y^{ts}_i , ||x^{ts}_i - \hat{x}^{ts}_i || \leq \epsilon\} }}{|\mathbf{D}^{ts}|} \\
    AUAC_{\epsilon_{max}}(f) = \frac{1}{\epsilon_{max}} \int_{\epsilon=0}^{\epsilon_{max}} Acc(f,\epsilon, \mathbf{D}^{ts}) d\epsilon.
\end{split}
\end{equation*}
\noindent
$Acc(f, \epsilon, \mathbf{D}^{ts})$ is the \emph{accuracy} of $f$ on the test set $\mathbf{D}^{ts}$ with perturbations of size up to $\epsilon$. $\epsilon_{max}$ is the highest perturbation considered for evaluation. In practice, to evaluate the accuracy for a given $\epsilon$, we count the number of perturbations found below that $\epsilon$.

Models must be defended against unknown attacks, so the evaluation cannot be fixed on a given $\epsilon$. It is important to know if the defended model will be resistant to all kinds of perturbation sizes. AUAC evaluates the performance trade-off for a range of perturbation sizes.
It takes into account the performance of a model from low to high perturbations. The closer the AUAC is to $1$, the more robust the classifier is. It offers a more accurate and fair way to compare defended models.

Having discussed how to compare and evaluate defenses, we have to find an upper bound $\epsilon_{max}$ for the perturbation size. In the following, we present our choice for this bound and our experimental results.

\section{Experiments} \label{sec:exp}

In this section we present an ablative study of the $\sigma$ parameter of our \emph{SAT} as well as a performance comparison with the state-of-the-art. First, we detail our experimental settings.

\subsection{Experimental settings} \label{sec:exp_settings}

We consider two popular datasets, namely, CIFAR-10 (6000 $32 \times 32$ RGB examples of each of 10 classes) and CIFAR-100 (600 $32 \times 32$ examples of each of 100 classes)~\cite{Krizhevsky2009LearningML}. For all experiments, the pixel values and perturbation size $\epsilon$ are normalized to [0,1] by dividing 255. We compare our \emph{SAT} to normal training as well as several state-of-the-art defenses:

\begin{itemize}
    \item \textbf{Standard Training (Standard)} trains with a cross entropy loss only on original training examples.
    \item \textbf{Madry Adversarial Training (Madry)}~\cite{madry_towards_2017} trains with a cross entropy loss only on adversarial examples generated at each iteration.
    \item \textbf{Mixed Adversarial Training (Mixed)}~\cite{goodfellow_explaining_2014}\cite{kurakin_adversarial_scale_2016} uses a mixed batch of adversarial examples (generated at each iteration) and original images with a cross entropy loss.
    \item \textbf{Adversarial Training with Domain Adaptation (ATDA)}~\cite{song_improving_2018} combines Mixed Adversarial Training with MMD and CORAL losses between adversarial and \\ original images. We do not include their Supervised Domain Adaptation loss since it does not lead to significant improvement according to their ablative study.
\end{itemize}


For each method, we reproduce their protocol and train a Resnet20~\cite{He_2016} or WideResnet 28-10 \cite{Zagoruyko_2016} on each dataset until convergence to provide a fair comparison with the same model architecture. 

To compute the Sinkhorn Divergences, we use the iterative Sinkhorn algorithm~\cite{Cuturi_2013} with 50 iterations.

For both models, we use a Stochastic Gradient Descent, with an initial learning rate of $0.1$, a weight decay of $5 \cdot 10^{-4}$ and a batch size of $128$.
For Resnet20, all the models converged after $60$ epochs of training with a multiplicative factor of $0.1$ on the learning rate at epoch $20$ and $40$.
For WideResnet 28-10, the models converged after $200$ epochs of training for Standard, Madry, Mixed and ATDA, and after 400 epochs for SAT. We add a multiplicative factor of $0.2$ on the learning rate at epoch $60$, $120$, $160$, $210$, $250$, $300$, $350$.



We perform $7$ iterations of $L_\infty$-PGD to generate our adversarial examples during training. For the evaluation, we apply $10$ iterations for PGD or $2000$ iterations for TR attack. All experiments are implemented on a single Titan X GPU.


Before evaluating the defenses, we have to define an upper bound $\epsilon_{max}$ on the perturbation size as mentioned on Section \ref{subsec:def_metric}.

\subsection{Choosing $\epsilon_{max}$}

\begin{figure}[hb]
\captionsetup[subfigure]{labelformat=empty}
  \begin{center}
    \subfloat[Original]{
      \includegraphics[width=0.1\textwidth]{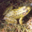}
                         }
    \subfloat[$\epsilon=16$]{
      \includegraphics[width=0.1\textwidth]{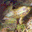}
                         }
    \subfloat[$\epsilon=30$]{
      \includegraphics[width=0.1\textwidth]{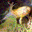}
                         }
    \hfill
    \subfloat[Original]{
      \includegraphics[width=0.1\textwidth]{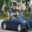}
                         }
    \subfloat[$\epsilon=16$]{
      \includegraphics[width=0.1\textwidth]{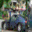}
                         }
    \subfloat[$\epsilon=30$]{
      \includegraphics[width=0.1\textwidth]{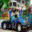}
                         }
    \caption{Examples of images from CIFAR-10 dataset perturbed with PGD at $\epsilon=16$ and $\epsilon=30$.}
    \label{fig:upperbound}
  \end{center}
\end{figure}

To compute our AUAC, we have to choose an upper bound $\epsilon_{max}$. This bound may depend on the format of images and datasets. In our case, we consider the same two possible choices for this bound on both datasets: $\epsilon_{max} = 16$ or $\epsilon_{max} = 30$.

As a first bound, we choose $\epsilon_{max}=16$ since beyond that the perturbation becomes visible to a human. For $\epsilon > 16$, we consider that the perturbation is out of the scope of adversarial examples, yet small enough for humans to recognize the images.


As a second bound, we choose $\epsilon_{max}=30$ since beyond that the class of the image becomes severely altered. In this case, for $\epsilon > 30$, a human would no longer recognize the class of the image. The perturbed images do not belong in their original cluster anymore.

Figure~\ref{fig:upperbound} gives examples from CIFAR-10 for both bounds. We can see that when $\epsilon=16$, the perturbation is barely detectable, but when $\epsilon=30$, the class of the image is severely altered and hard to recognize.

With the upper bounds defined, we can evaluate and compare defenses for both of them.

\subsection{Ablative study on the entropy parameter} \label{sec:sigma_abl}

Given that the properties of Sinkhorn Divergences vary depending on the value of the entropy parameter $\sigma$, we examine its impact on robustness. 

\begin{figure}[t]
    \centering
    \includegraphics[width=\columnwidth]{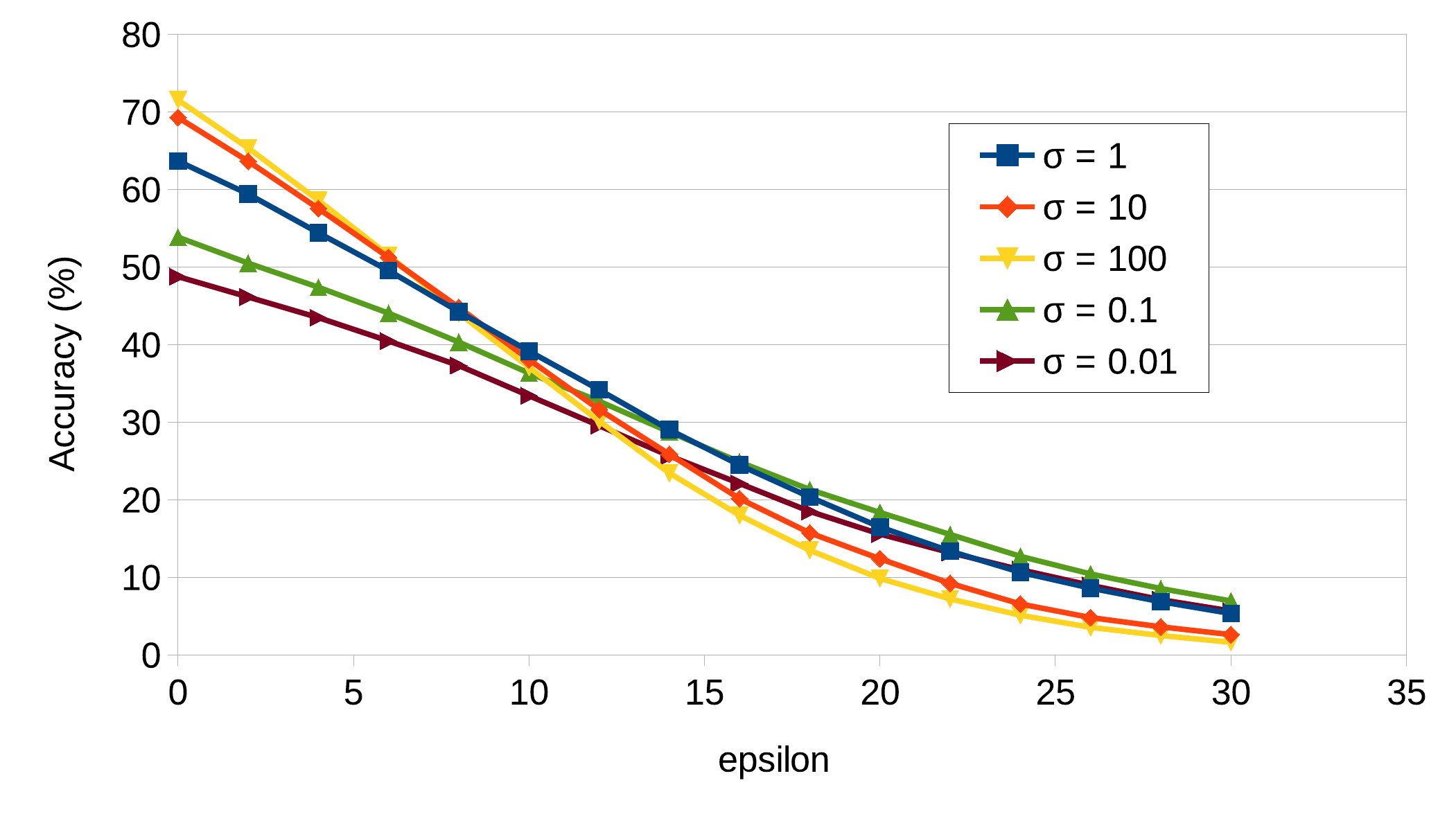}
    \caption{Accuracy of SAT on CIFAR-10 for different entropy $\sigma$ against adversarial examples of varying size $\epsilon$ generated with PGD. Defenses use a Resnet20 and $\epsilon_{train}=8$.}
    \label{fig:SAT_abl_resnet}
\end{figure}

\begin{figure}[t]
    \centering
    \includegraphics[width=\columnwidth]{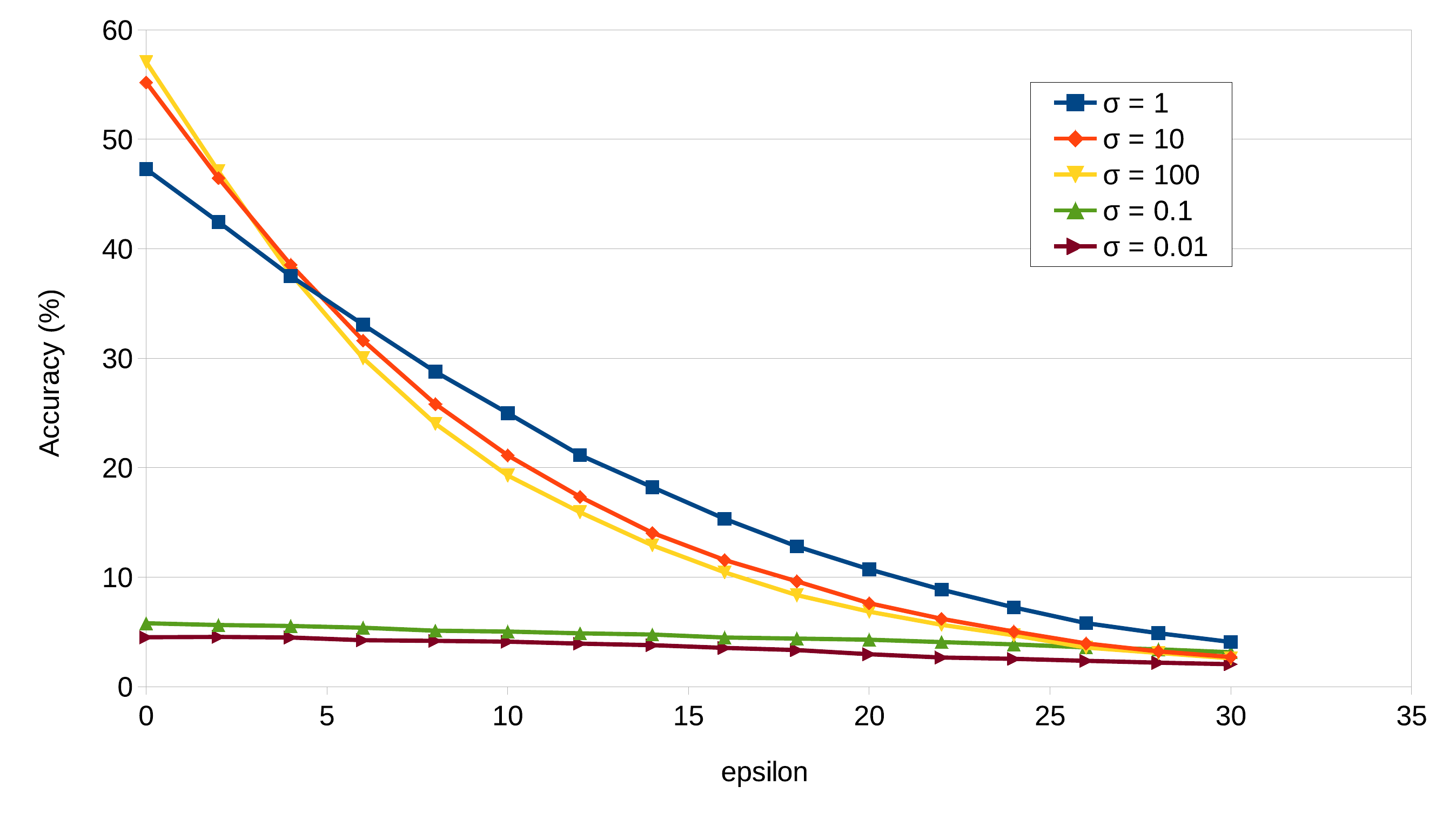}
    \caption{Accuracy of SAT on CIFAR-100 for different entropy $\sigma$ against adversarial examples of varying size $\epsilon$ generated with PGD. Defenses use a WideResnet28-10 and $\epsilon_{train}=8$.}
    \label{fig:SAT_abl_wide}
\end{figure}

Figures~\ref{fig:SAT_abl_resnet} and \ref{fig:SAT_abl_wide} compare the accuracy of SAT for different values of $\sigma$ and perturbation size $\epsilon$ on CIFAR-10 and CIFAR-100. What can clearly be seen in these figures is that the performance depends on the accuracy on the original images, and how the accuracy decreases when $\epsilon$ increases. The original accuracy gives a good approximation of the performance for small $\epsilon$, but not for high perturbations. Furthermore, comparing the performance with single $\epsilon$ does not give a good overview of the robustness of the defense. For instance, on CIFAR-100, for $\epsilon = 2$, $\text{SAT}_1$ and $\text{SAT}_{100}$ have respectively $42.46\%$ and $47.09\%$ accuracy but for $\epsilon = 8$, the same models have respectively $28.77\%$ and $24.02\%$ accuracy. $\text{SAT}_1$ has the best performance when $\epsilon > 5$ whereas $\text{SAT}_{100}$ is more robust when $\epsilon < 5$. This emphasizes the need for a more accurate evaluation metric for robustness.

Table~\ref{tab:SAT_auc} provides AUAC up to $\epsilon_{max}=16$ or $\epsilon_{max}=30$ depending on the entropy parameter $\sigma$ for Resnet20 on CIFAR-10 and WideResnet28-10 on CIFAR-100. We can see that SAT is the most robust overall when $\sigma=1$. For instance, on CIFAR-100 with $\sigma=1$, SAT has an AUAC of $29.69 \%$ up to $\epsilon_{max}=16$ and $19.83\%$ when $\epsilon_{max}=30$. 

In summary, these results show that using $\sigma=1$ results in a more robust defense over a wide range of perturbation sizes. This entropy represents a hinge value between an MMD and an Optimal Transport loss~\cite{genevay:hal-02411822}. In the following, we will always consider SAT with an entropy of $1$.

Let us now compare the performance of SAT with other state-of-the-art defenses.


\begin{table}[t]
    \begin{center}
    \begin{tabular}{@{}llll@{}} \toprule
    Dataset / Architecture & $\sigma$ & AUAC@16 (\%) & AUAC@30 (\%) \\
    \midrule
    \multirow{5}{*}{CIFAR-10 / Resnet20} & 1 & $44.26$ & $\mathbf{29.69}$ \\
     & 10 & $\mathbf{44.69}$ & $28.09$ \\
     & 100 & $44.38$ & $27.11$ \\
     & 0.1 & $39.94$ & $28.16$ \\
     & 0.01 & $36.44$ & $25.32$ \\
     \midrule
    \multirow{5}{*}{CIFAR-100 / WideResnet28-10} & 1 & $\mathbf{29.69}$ & $\mathbf{19.83}$ \\
     & 10 & $28.55$ & $18.08$ \\
     & 100 & $27.59$ & $17.30$ \\
     & 0.1 & $5.18$ & $4.59$ \\
     & 0.01 & $4.16$ & $3.48$ \\
    \bottomrule
    \end{tabular}
    \end{center}
    \caption{Comparison of AUAC up to $\epsilon_{max}=16$ and $\epsilon_{max}=30$ against PGD for different values of $\sigma$ in SAT.}
    \label{tab:SAT_auc}
\end{table}


\subsection{Comparison with other defenses}

In the following experiments, we compare the performance of our SAT (with $\sigma = 1$) to other state-of-the-art defenses, first against PGD then against TR attack.

\begin{figure}
    \centering
    \includegraphics[width=\columnwidth]{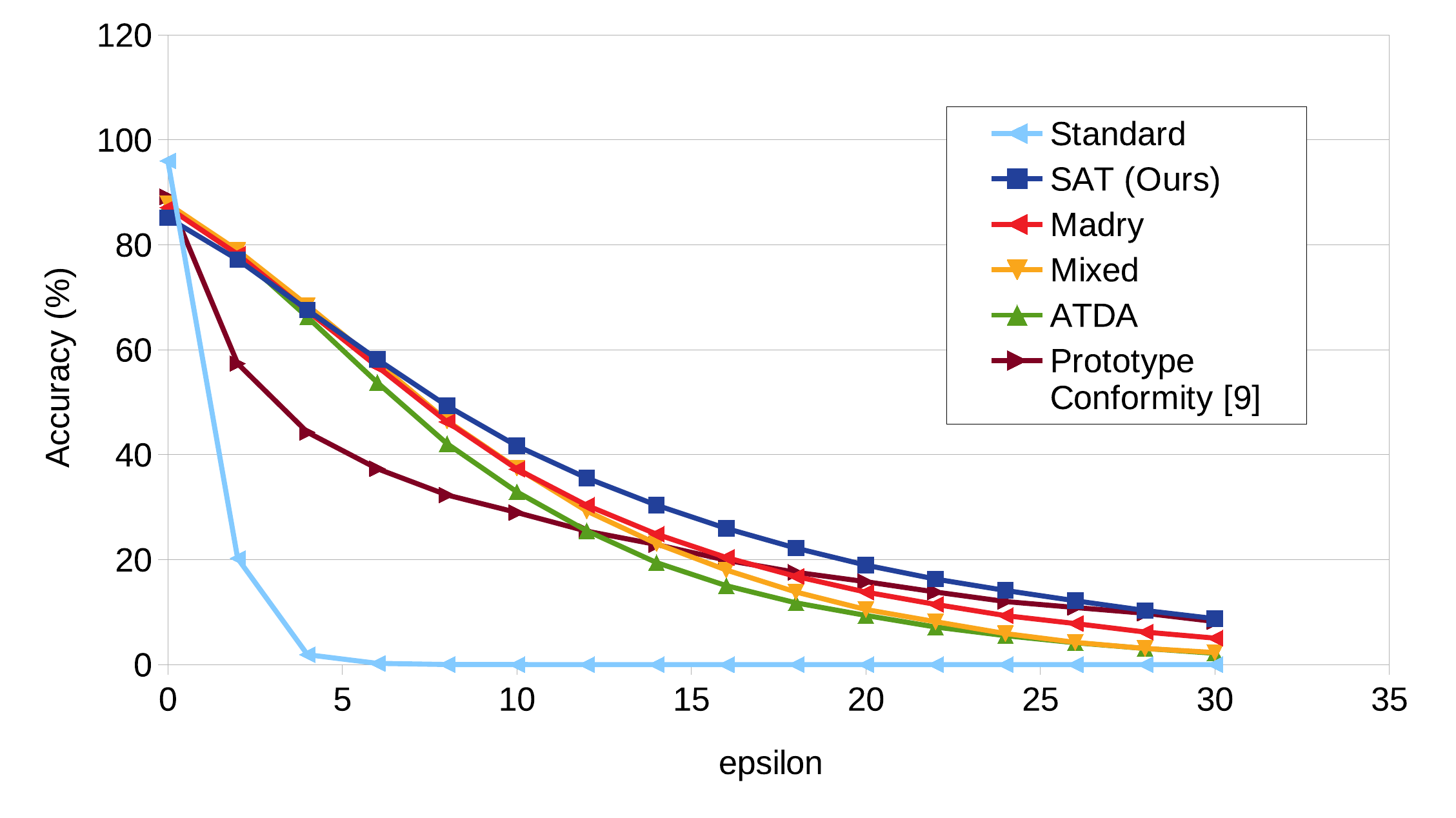}
    \caption{Performance of the defenses on CIFAR-10 against PGD for perturbation sizes up to $\epsilon=30$. Our SAT and the other reproduced defenses use a WideResnet28-10 and adversarial examples generated with $\epsilon_{train}=8$. Prototype Conformity defense uses a Resnet110 with a uniformly sampled $\epsilon_{train}$ between $3$ and $13$.}
    \label{fig:sota_comp_cifar10}
\end{figure}

Figure~\ref{fig:sota_comp_cifar10} provides an overview of the accuracy on CIFAR-10 with $\epsilon_{train}=8$ against PGD over a wide range of perturbation sizes, from $\epsilon=0$ (original accuracy) to $\epsilon=30$. In addition to the methods presented in Section~\ref{sec:exp_settings}, we report the results for the Prototype Conformity~\cite{Mustafa_2019_ICCV} (PC) using the weights publicly available.
Interestingly, this defense is less robust to low perturbations than other methods. 
On the other hand, other defenses have similar performance at low perturbations but they do not react in the same way when the size of the perturbation increases. ATDA has the lowest performance when $\epsilon \geq 12$ and SAT is the most robust method overall.
For instance, for $\epsilon=8$, our SAT has an accuracy of $49.31\%$ whereas Mixed has $46.46\%$, Madry $46.26\%$, ATDA $42.1\%$, PC $32.32\%$ and the Standard method $0.02\%$.

\begin{figure}
    \centering
    \includegraphics[width=\columnwidth]{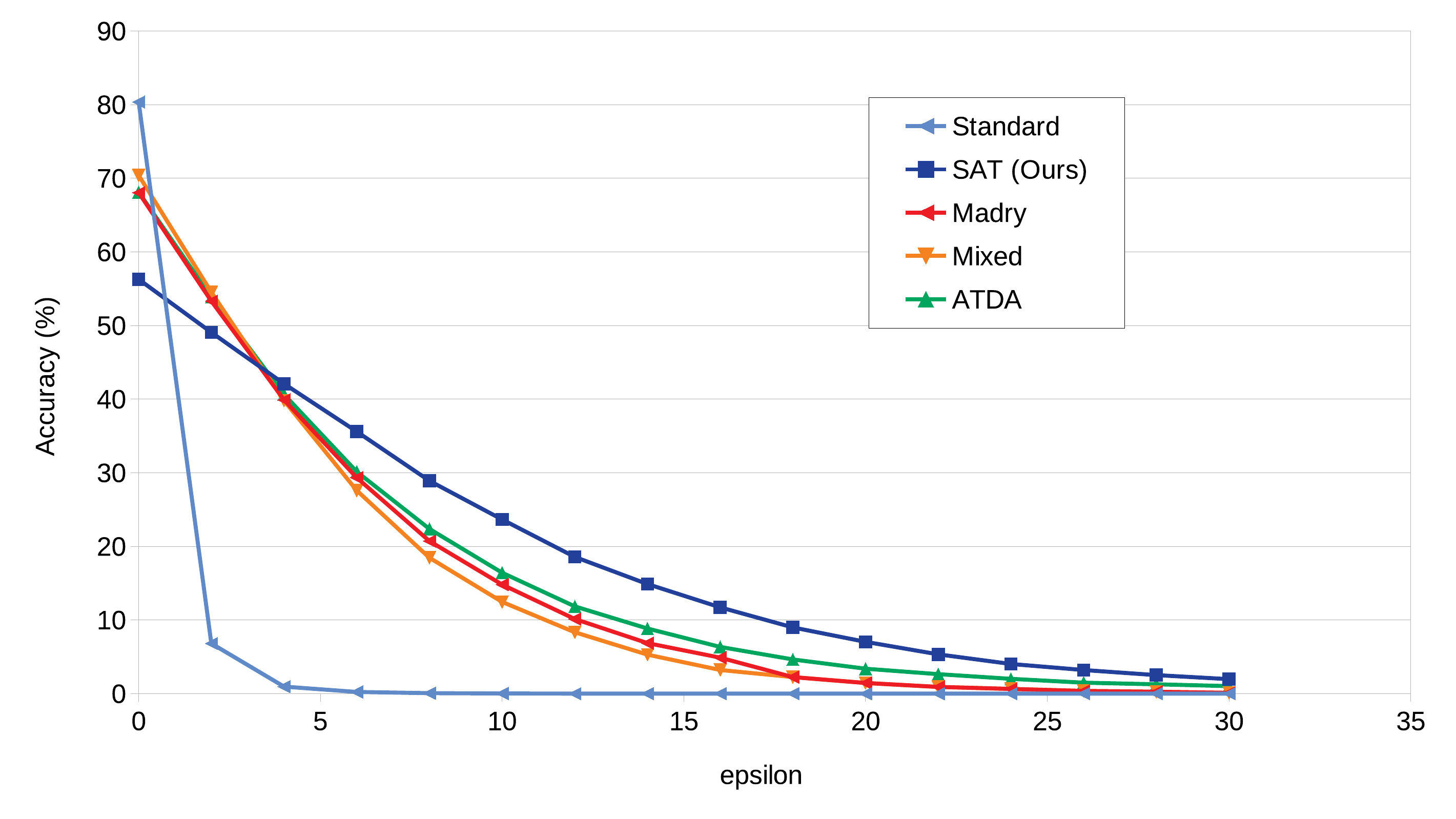}
    \caption{Performance of the defenses on CIFAR-100 against PGD for perturbation sizes up to $\epsilon=30$. Our SAT and the other reproduced defenses use a WideResnet28-10 and adversarial examples generated with $\epsilon_{train}=4$.}
    \label{fig:sota_comp_cifar100}
\end{figure}

Similarly, Figure~\ref{fig:sota_comp_cifar100} presents a comparison of the performance on CIFAR-100 against PGD for the same range of perturbation size but with $\epsilon_{train}=4$.
In this case, it can be seen that all defenses except SAT have very similar performance. SAT is significantly more robust to large perturbations than the other methods but not as robust to small perturbations. SAT achieves better robustness overall but with lower original accuracy. For instance, for $\epsilon=4$, our SAT has an accuracy of $42.09\%$, whereas Mixed has $39.8\%$, Madry $39.91\%$, ATDA $40.64\%$ and the Standard method $0.95\%$.

The results of the AUAC against PGD on both datasets are provided in Table~\ref{tab:comp_AUAC_pgd}. SAT is the most robust defense over all perturbations used during evaluation. For instance, on CIFAR-10 with $\epsilon_{train} = 8$, SAT has an AUAC up to $\epsilon=16$ of $51.93\%$ whereas ATDA has $46.19\%$.

\begin{figure}
    \centering
    \includegraphics[width=\columnwidth]{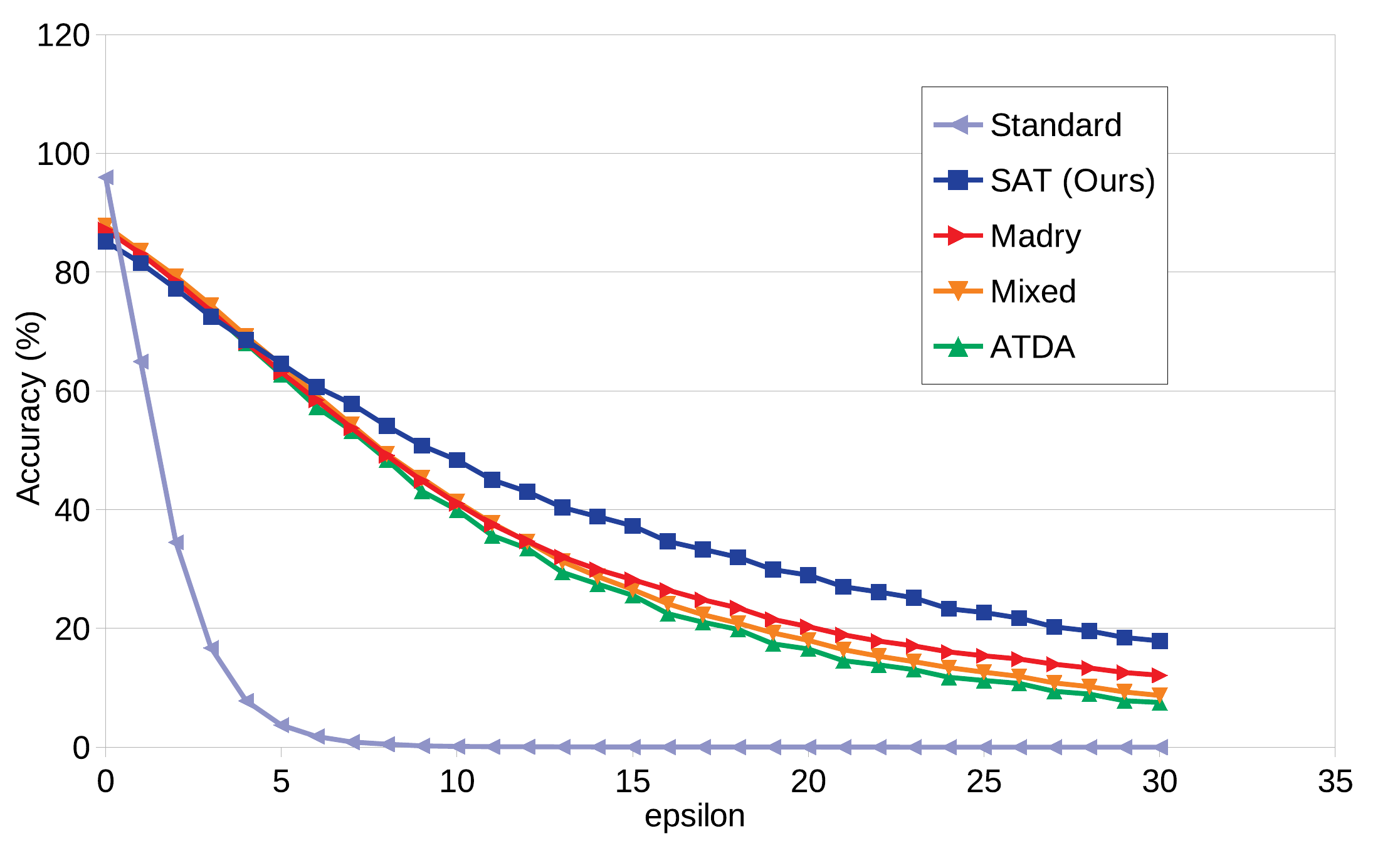}
    \caption{Performance of the defenses on CIFAR-10 against TR attack for perturbation sizes up to $\epsilon=30$. Our SAT and the other reproduced defenses use a WideResnet28-10 and adversarial examples generated with $\epsilon_{train}=8$.}
    \label{fig:sota_comp_cifar10_tr}
\end{figure}

\begin{figure}
    \centering
    \includegraphics[width=\columnwidth]{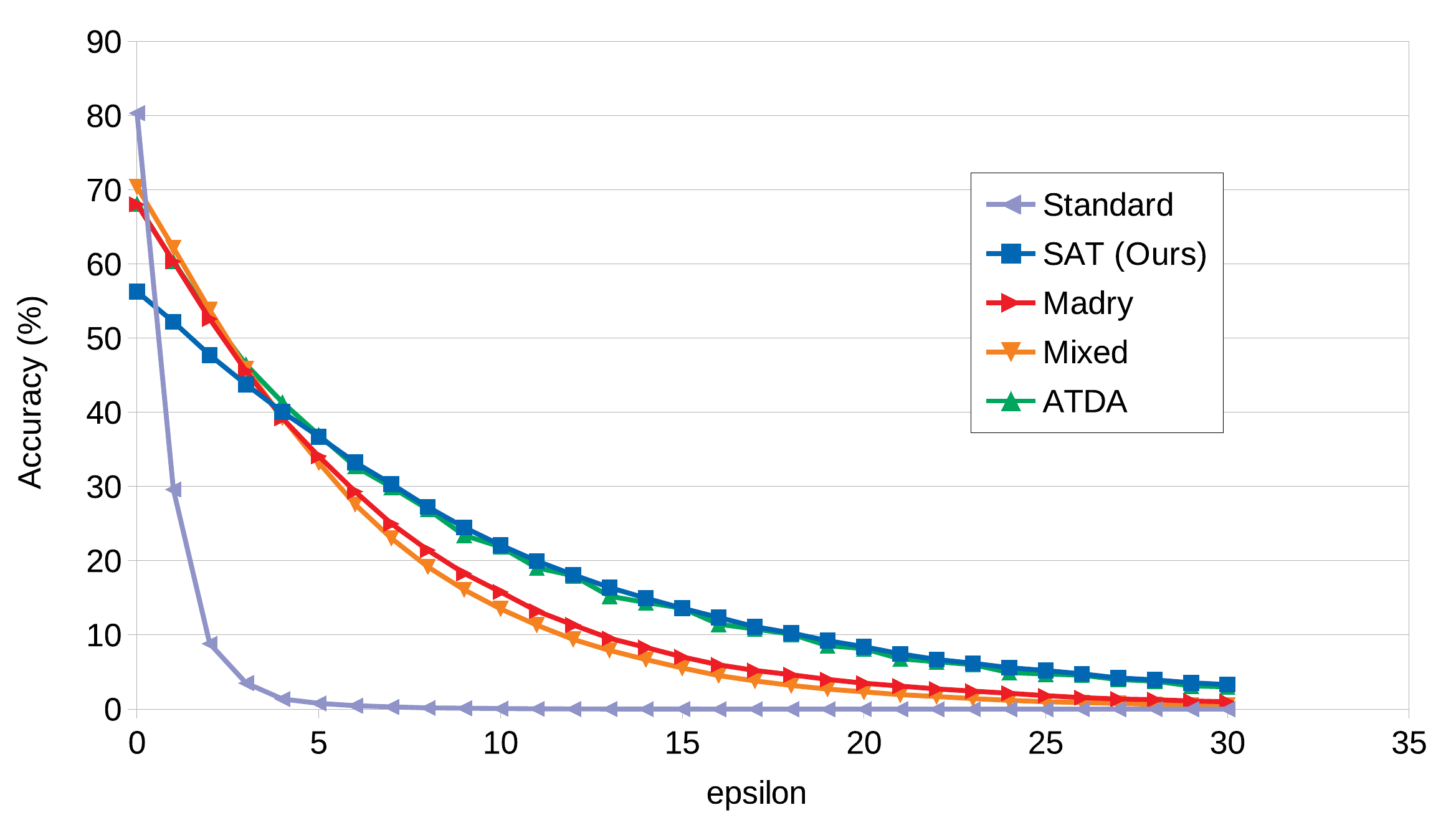}
    \caption{Performance of the defenses on CIFAR-100 against TR attack for perturbation sizes up to $\epsilon=30$. Our SAT and the other reproduced defenses use a WideResnet28-10 and adversarial examples generated with $\epsilon_{train}=4$.}
    \label{fig:sota_comp_cifar100_tr}
\end{figure}

Figures~\ref{fig:sota_comp_cifar10_tr} and \ref{fig:sota_comp_cifar100_tr} compares the performance of the defenses against TR attack.
On CIFAR-10, we can see that all the defenses have similar performance against small perturbations ($\epsilon \leq 5$) but SAT is significantly more robust to higher perturbations. 
On CIFAR-100, SAT has lower performance on small perturbations than the other defenses. However, SAT and ATDA are the most robust to high perturbations.

Table~\ref{tab:comp_AUAC_tr} provides the AUAC against TR attack on both datasets when training on adversarial examples generated with $\epsilon_{train}=4$ or $\epsilon_{train}=8$. On CIFAR-10, our SAT is significantly more robust than the other defenses. On CIFAR-100, despite good robustness on high perturbations, the lower accuracy on small perturbations leads to a lower AUAC for SAT than ATDA. However, SAT still has competitive AUAC compared to other defenses.

Note that when increasing $\epsilon_{train}$, the defended models become more robust to high perturbations but the accuracy on original images decreases. However, the accuracy on high perturbations is also bounded by the accuracy on original images. Thus, the AUAC can help finding a good trade-off on $\epsilon_{train}$ between accuracy on original images and robustness to high perturbations. 

In this section, we provided an experimental analysis of our SAT. First, we presented an ablative study on the entropy parameter $\sigma$ which resulted in fixing $\sigma=1$. It can be interpreted as using a Sinkhorn Divergence at the boundary between an MMD loss and an Optimal Transport loss. Then, we compared our SAT to other state-of-the-art defenses and showed that it is a more robust defense on a wide range of perturbation sizes.

\begin{table}[t]
    \begin{center}
    \begin{tabular}{@{}lllll@{}} \toprule
    Dataset & Archi. & Model & \multicolumn{2}{c}{AUAC (\%)} \\
    \cmidrule{4-5}
     & & & $\epsilon_{max}=16$ & $\epsilon_{max} = 30$ \\
    \midrule
    \multirow{9}{*}{CIFAR-10} & \multirow{4}{*}{Resnet20} & Standard & $5.79$ & $3.09$ \\
     & & Madry & $44.18$ & $26.53$ \\
     & & Mixed & $40.68$ & $22.73$ \\
     & & ATDA & $35.58$ & $21.63$ \\
     & & SAT (Ours) & $\mathbf{44.26}$ & $\mathbf{29.69}$ \\
    \cmidrule{2-5}
    & Resnet110 & PC \cite{Mustafa_2019_ICCV} & $37.89$ & $26.47$ \\
    \cmidrule{2-5}
     & \multirow{4}{*}{WideResnet28-10} & Standard & $8.8$ & $4.69$ \\
     & & Madry & $49.37$ & $31.54$ \\
     & & Mixed & $49.27$ & $30.01$ \\
     & & ATDA & $46.19$ & $27.94$ \\
     & & SAT (Ours) & $\mathbf{51.93}$ & $\mathbf{35.12}$ \\
    \midrule
    \multirow{4}{*}{CIFAR-100} & \multirow{4}{*}{WideResnet28-10} & Standard & $6.03$ & $3.22$ \\
     & & Madry & $27.27$ & $16.14$ \\
     & & Mixed & $27.80$ & $16.13$ \\
     & & ATDA & $28.59$ & $17.11$ \\
     & & SAT (Ours) & $\mathbf{29.69}$ & $\mathbf{19.83}$ \\
    \bottomrule
    \end{tabular}
    \end{center}
    \caption{Comparison of AUAC (in \%) up to $\epsilon_{max}=16$ or $\epsilon_{max}=30$ against PGD on both CIFAR-10 and CIFAR-100. Our SAT and other defenses reproduced are trained with adversarial examples generated with $\epsilon_{train}=8$.}
    \label{tab:comp_AUAC_pgd}
\end{table}

\begin{table*}[t]
    \begin{center}
    \begin{tabular}{@{}lllllll@{}} \toprule
    \multirow{3}{*}{Dataset} & \multirow{3}{*}{Archi.} & \multirow{3}{*}{Model} & \multicolumn{4}{c}{AUAC (\%)} \\
    \cmidrule{4-7}
     & & & \multicolumn{2}{c}{$\epsilon_{max} = 16$} & \multicolumn{2}{c}{$\epsilon_{max} = 30$} \\
     \cmidrule{4-7}
     & & & $\epsilon_{train} = 4$ & $\epsilon_{train} = 8$ & $\epsilon_{train} = 4$ & $\epsilon_{train} = 8$   \\
    \midrule
    \multirow{4}{*}{CIFAR-10} & \multirow{4}{*}{WideResnet28-10} & Standard & \multicolumn{2}{c}{$11.23$} & \multicolumn{2}{c}{$6.00$} \\
     & & Madry &  $52.56$ & $52.05$ & $34.93$ & $36.07$ \\
     & & Mixed & $51.50$  & $52.22$ & $33.44$ & $34.89$\\
     & & ATDA & $49.43$ & $50.94$ & $31.04$ & $33.54$\\
     & & SAT (Ours) & $\mathbf{54.80}$ & $\mathbf{56.29}$ & $\mathbf{35.99}$ & $\mathbf{41.85}$\\
    \midrule
    \multirow{4}{*}{CIFAR-100} & \multirow{4}{*}{WideResnet28-10} & Standard & \multicolumn{2}{c}{$5.34$} & \multicolumn{2}{c}{$2.85$} \\
     & & Madry & $26.76$ & $27.25$ & $15.55$ & $16.64$\\
     & & Mixed & $25.76$ & $27.95$ & $14.56$ & $16.71$\\
     & & ATDA & $\mathbf{30.82}$ & $\mathbf{30.15}$ & $\mathbf{19.40}$ & $\mathbf{19.64}$\\
     & & SAT (Ours) & $29.71$ & $27.80$ & $18.99$ & $19.27$ \\
    \bottomrule
    \end{tabular}
    \end{center}
    \caption{Comparison of AUAC (in \%) up to $\epsilon_{max}=16$ or $\epsilon_{max}=30$ against TR attack on both CIFAR-10 and CIFAR-100. Defenses are trained with adversarial examples generated with $\epsilon_{train}=4$ or $\epsilon_{train}=8$.}
    \label{tab:comp_AUAC_tr}
\end{table*}

\section{Conclusion}


Defending against adversarial attacks is a critical issue for practical applications using deep learning. Images can easily be altered in such a way that undefended models make mistakes, which can lead to security breaches.

In this work, we follow the analogy with domain adaptation and consider separate distributions of original and adversarial images. We aim to improve the robustness of models to adversarial attacks by aligning the representations of both distributions using a least effort approach. 
In addition, we raise concerns on the recurring trend to evaluate defenses with a single perturbation size similar to the one use during training. It does not reflect the variations in robustness depending on the size of the perturbations.

First, we propose SAT (Sinkhorn Adversarial Training), a new defense using the theory of optimal transport with Sinkhorn Divergences. We minimize the ground distance between representations to bring closer the distributions and take into account the geometry of space.
Then, for a fair evaluation of robustness, we propose the Area Under the Accuracy Curve (AUAC) to compare defenses. The AUAC integrates the performance on a wide range of perturbation sizes to quantify robustness.

Finally, we perform a thorough analysis on CIFAR-10 and CIFAR-100 datasets and show that SAT outperforms other defenses by comparing accuracy curves and our AUAC. 






\bibliographystyle{IEEEtran}
\bibliography{references.bib}
%

\end{document}